\pdfoutput=1

\documentclass[11pt]{article}

\usepackage[preprint]{acl}

\usepackage{times}
\usepackage{latexsym}

\usepackage[T1]{fontenc}

\usepackage[utf8]{inputenc}

\usepackage{microtype}

\usepackage{inconsolata}

\usepackage{stfloats}
\usepackage{colortbl, xcolor}
\usepackage{graphicx}
\usepackage{subcaption}
\usepackage{longtable}
\usepackage{soul, color, xcolor}
\usepackage{booktabs}
\usepackage{xspace}

\usepackage{array}
\usepackage{ragged2e}
\usepackage{enumitem}
\usepackage{pifont}
\usepackage{tikz}
\usepackage{xcolor}
\usepackage{graphicx}
\usepackage{xspace}
\usepackage{stfloats} %

\usepackage[most]{tcolorbox}
\usepackage{setspace} %
\definecolor{customblue}{RGB}{25, 18, 180}

\usepackage{hyperref}
\hypersetup{
  colorlinks   = true,    %
  linkcolor    = customblue,    %
  breaklinks   = true,
}

\usepackage[nameinlink, capitalise]{cleveref}       %

\usepackage{cancel}
\usepackage{bigdelim, rotating}  %

\newcommand{\eg}{\emph{e.g}.\xspace}
\newcommand{\ie}{\emph{i.e}.\xspace}

\usetikzlibrary{tikzmark}
\makeatletter
\newcommand*\myfontsize{%
\@setfontsize\myfontsize{6.7}{8}%
}
\makeatother

\definecolor{cadmiumgreen}{rgb}{0.0, 0.42, 0.24}

\definecolor{myred}{rgb}{0.7, 0.3, 0.0}
\definecolor{myblue}{rgb}{0.2, 0.3, 0.6}

\newenvironment{packeditemize}{
\begin{list}{$\bullet$}{
\setlength{\labelwidth}{6pt}
\setlength{\itemsep}{0pt}
\setlength{\leftmargin}{\labelwidth}
\addtolength{\leftmargin}{\labelsep}
\setlength{\parindent}{0pt}
\setlength{\listparindent}{\parindent}
\setlength{\parsep}{0pt}
\setlength{\topsep}{3pt}}}{\end{list}}

\renewcommand{\texttt}[1]{%
  \begingroup
  \ttfamily
  \begingroup\lccode`~=`/\lowercase{\endgroup\def~}{/\discretionary{}{}{}}%
  \begingroup\lccode`~=`[\lowercase{\endgroup\def~}{[\discretionary{}{}{}}%
  \begingroup\lccode`~=`.\lowercase{\endgroup\def~}{.\discretionary{}{}{}}%
  \begingroup\lccode`~=`-\lowercase{\endgroup\def~}{-\discretionary{}{}{}}%
  \begingroup\lccode`~=`]\lowercase{\endgroup\def~}{]\discretionary{}{}{}}%
  \begingroup\lccode`~=`_\lowercase{\endgroup\def~}{\_\discretionary{}{}{}}%
  \catcode`/=\active
  \catcode`[=\active
  \catcode`.=\active
  \catcode`-=\active
  \catcode`]=\active
  \catcode`_=\active
  \scantokens{#1\noexpand}%
  \endgroup
}

\newcommand{\llm}[1]{\texttt{#1}}
\newcommand{\dataset}[1]{\texttt{#1}}

\newcolumntype{L}[1]{>{\raggedright\arraybackslash}p{#1}}
\newcolumntype{C}[1]{>{\centering\arraybackslash}p{#1}}
\newcolumntype{R}[1]{>{\raggedleft\arraybackslash}p{#1}}
\newcolumntype{M}[1]{>{\centering\arraybackslash}m{#1}}
\newcolumntype{P}[1]{>{\raggedright\arraybackslash}m{#1}}

\NewEnviron{promptgroup}{%
  \begin{minipage}{\columnwidth}
    \BODY
  \end{minipage}
}

\newcounter{sharedbox}                   %
\crefname  {sharedbox}{Prompt}{Prompts}  %
\Crefname  {sharedbox}{Prompt}{Prompts}  %

\newtcolorbox{promptbox}[2][]{%
  float,                    %
  floatplacement=!t,      %
  width=\linewidth,         %
  colback=white,            %
  colframe=black,           %
  coltitle=black,           %
  colbacktitle=white,       %
  boxrule=1.2pt,
  left=5pt, right=5pt, top=5pt, bottom=5pt,
  before upper={\setstretch{0.8}},  %
  fonttitle=\sffamily\bfseries\small,
  title={%
    \refstepcounter{sharedbox}%
    \label{#1}%
    \fontsize{9.7}{7}\selectfont Prompt \thesharedbox: #2%
  },%
}

\newtcolorbox{promptboxc}[2][]{%
  float*=!t,                 %
  width=\textwidth,
  colback=white,
  colframe=black,
  coltitle=black,
  colbacktitle=white,
  boxrule=1.2pt,
  left=5pt, right=5pt, top=5pt, bottom=5pt,
  before upper={\setstretch{0.8}}, 
  fonttitle=\sffamily\bfseries\small,
  title={%
    \refstepcounter{sharedbox}%
    \label{#1}%
    \fontsize{9.7}{7}\selectfont Prompt \thesharedbox: #2%
  },%
}

\newtcolorbox[auto counter]{examplebox}[2][]{
  float,
  float=htbp,  %
  width=\linewidth,
  colback=white,
  title={\fontsize{9.7}{7}\selectfont Example \thetcbcounter: #2},
  coltitle=black,
  left=5pt,
  right=5pt,
  top=5pt,
  bottom=5pt,
  fonttitle=\sffamily\bfseries\small,
  boxrule=1.2pt,
  label={#1},
  colframe=black,
  colbacktitle=white,
  before upper={\setstretch{0.9}},
  before={\par\vspace*{0pt}},
  after={\par\vspace*{0pt}},
}

\crefname{figure}{Figure}{Figures}
\Crefname{figure}{Figure}{Figures}
\crefname{table}{Table}{Tables}
\Crefname{table}{Table}{Tables}
\crefname{equation}{Eq.}{Eqs.}
\Crefname{equation}{Eq.}{Eqs.}
\crefname{appendix}{Appendix}{Appendices}
\Crefname{appendix}{Appendix}{Appendices}

\makeatletter
\crefname{tcb@cnt@examplebox}{example}{examples}    %
\Crefname{tcb@cnt@examplebox}{Example}{Examples}    %
\makeatother

\usepackage{arydshln} %

\newcommand{\ourbench}{\textbf{CogTest}\xspace}
\newcommand{\habit}[1]{\textit{#1}\xspace}

\title{Towards Understanding the Cognitive Habits of Large Reasoning Models}

\author{Jianshuo Dong$^{1}$, Yujia Fu$^{1}$, Chuanrui Hu$^{2}$, Chao Zhang$^{1}$, Han Qiu$^{1}$ \\
$^{1}$Tsinghua University, China. $^{2}$Qihoo 360, China.\\
\texttt{Emails: dongjs23@mails.tsinghua.edu.cn}
}

\begin{document}
\maketitle
\begin{abstract}
Large Reasoning Models (LRMs), which autonomously produce a reasoning Chain of Thought (CoT) before producing final responses, offer a promising approach to interpreting and monitoring model behaviors.
Inspired by the observation that certain CoT patterns—\eg, \textit{Wait, did I miss anything?}—consistently emerge across tasks, we explore whether LRMs exhibit human-like cognitive habits.
Building on \textit{Habits of Mind}, a well-established framework of cognitive habits associated with successful human problem-solving, we introduce \ourbench, a principled benchmark designed to evaluate LRMs' cognitive habits.
\ourbench includes 16 cognitive habits, each instantiated with 25 diverse tasks, and employs an evidence-first extraction method to ensure reliable habit identification. 
With \ourbench, we conduct a comprehensive evaluation of 16 widely used LLMs (13 LRMs and 3 non-reasoning ones).
Our findings reveal that LRMs, unlike conventional LLMs, not only exhibit human-like habits but also adaptively deploy them according to different tasks
Finer-grained analyses further uncover patterns of similarity and difference in LRMs' cognitive habit profiles, particularly certain inter-family similarity (\eg, \llm{Qwen-3} models and \llm{DeepSeek-R1}).
Extending the study to safety-related tasks, we observe that certain habits, such as \habit{Taking Responsible Risks}, are strongly associated with the generation of harmful responses.
These findings suggest that studying persistent behavioral patterns in LRMs' CoTs is a valuable step toward deeper understanding of LLM misbehavior.
The code is available at: \url{https://github.com/jianshuod/CogTest}.
\end{abstract}

\section{Introduction}

Large Reasoning Models (LRMs), including OpenAI o1~\citep{jaech2024openai-o1-system-card}, DeepSeek-R1~\citep{guo2025deepseek-r1}, and Gemini-thinking models~\citep{gemini-2.5-thinking}, have recently garnered significant attention. 
Unlike non-reasoning models, LRMs autonomously generate a chain of thought (CoT) before producing a final response.
This ability, often acquired through reinforcement learning or distillation~\citep{jaech2024openai-o1-system-card, guo2025deepseek-r1,muennighoff2025s1}, substantially enhances the reasoning capabilities of large language models.
However, CoT reasoning also introduces new risks, with LRMs exhibiting problematic behaviors ranging from overthinking~\citep{chen2024not-tencent-llm-overthinking} to increased safety concerns~\citep{zhang2025should-safety-of-lrms}.

Meanwhile, the thinking-then-answering paradigm provides a valuable lens for interpreting and monitoring the rationales behind LRM responses~\citep{openai2025misbehavior-cot-monitoring,chen2025reasoning-anthropic-unfaithful-cot-for-monitoring}.
CoTs typically reveal how LRMs process given instructions and how they progress towards final responses.
For example, an observation of the \llm{DeepSeek-R1}'s reasoning CoTs reveals reflective thinking pattern: ``Wait, did I miss anything?''
These patterns resemble human cognitive behaviors during problem-solving and appear consistently across varying instructions, independent of specific tasks.
This invites an intriguing research question:
\textit{Do LRMs exhibit human-like ``cognitive habits'' that underpin their strong problem-solving abilities? }

To answer this, we adapt the \textit{Habits of Mind} framework~\citep{costa2005habits-of-mind-book} to systematically examine whether the cognitive habits commonly observed in successful human problem-solving are exhibited by LRMs as well.
This framework comprises 16 positive problem-solving habits, such as \habit{thinking about thinking} and \habit{managing impulsivity}.
Building on the framework, our testing of LRM cognitive habits follows a three-stage process:
First, we curate high-quality tasks tailored to each habit.
Next, we elicit the reasoning CoTs of LRMs in solving the tasks.
Finally, we determine whether the target cognitive habit is exhibited in the reasoning CoT produced by the LRM.

\begin{figure*}[t]
    \centering
    \includegraphics[width=1\linewidth]{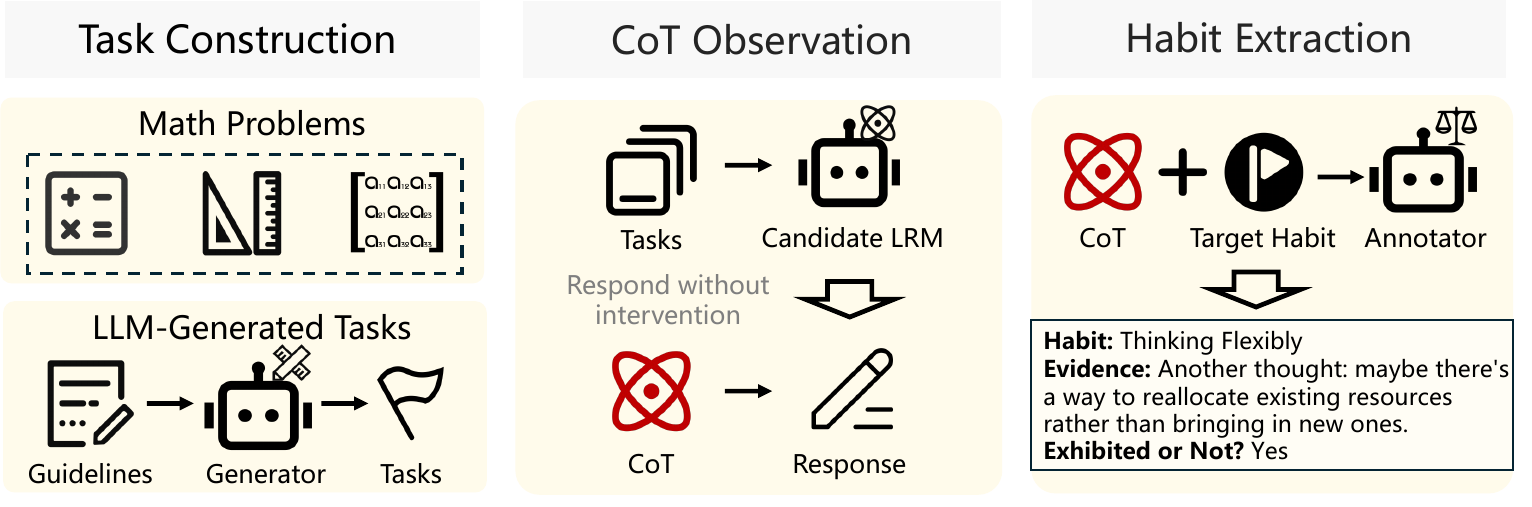}
    \caption{\textbf{Pipeline of measuring cognitive habits of LRMs}. 
    }
    \label{fig:pipeline}
\end{figure*}

Operationally, we instantiate each habit with 25 tasks that can effectively differentiate the LRMs' inherent possession of specific cognitive habits (\textit{Habit Specificity and Comprehensiveness}).
These tasks are carefully designed to reflect the nature of each habit while avoiding any explicit mention or implicit cues about the targeted habit (\textit{Spontaneity}).
Furthermore, the tasks are grounded in real-world scenarios to better capture how LRMs react to complex tasks (\textit{Real-World Utility}).
The task construction is achieved in a hybrid way:
For thinking-related habits, we recognize that math problems are an ideal testbed and employ MATH-500~\citep{hendrycks2measuring-math-500} as our task source.
For the remaining habits, we start with human-designed guidelines for each habit and prompt advanced LLMs (\ie, \llm{GPT-4.1}) for automated task generation.
We introduce a benchmark, \ourbench, consisting of 25 carefully designed habit-inducing tasks for each cognitive habit.
As math problems are widely available and habit-specific guidelines can be reused, the pipeline enjoys the benefits of \textit{scalability}.
We make huge efforts in manual verification to ensure the quality of the \ourbench benchmark.
Along with the tasks, we also establish an automatic evidence-first habit extraction method, which employs LLM-as-a-judge~\citep{zheng2024llm-judge-fastchat} and requests a reference from CoT as evidence before making the judgment.
This enables effective identification of habits exhibited in LLMs' reasoning CoTs and minimizes the risk of hallucination~\citep{zhang2023siren-song-survey-hallucination-llms}.

In this work, we give initial evidence that LRMs indeed exhibit cognitive habits and certain LRMs like \llm{DeepSeek-R1} are highly capable of demonstrating suitable habits according to the surfaced tasks.
Our comprehensive testing covers 16 renowned LLMs, including 13 LRMs and 3 non-reasoning LLMs prompted to generate explicit reasoning CoTs.
Comparative analysis reveals notable differences between reasoning and non-reasoning models: 
Non-reasoning LLMs, although exhibiting habits like \habit{thinking and communicating with clarity and precision}, struggle with generating extended CoTs.
This reveals the importance of reasoning RL in boosting LRMs' reasoning abilities.
We observe clear similarities among intra-family and post-distillation models, whereas inter-family models such as \llm{Claude-3.7-sonnet} versus others exhibit distinct cognitive habit profiles.
Interestingly, \llm{DeepSeek-R1} and \llm{Qwen-3} exhibit notable resemblance despite originating from different families.

Moreover, we extend our analysis to 200 safety-related queries from~\citet{mazeika2024harmbench-automated-red-teaming}. 
Our results suggest that the presence of certain cognitive habits can correlate with LRMs' harmful responses. 
For example, the habit \habit{Taking Responsible Risks} is associated with a higher incidence of harmful responses, indicating that some LRMs may recognize potential risks yet proceed to engage with harmful prompts. 
This underscores the need for a deeper understanding of the cognitive habits implicit in CoTs as a means to monitor and mitigate model risks, in line with the proposal by~\citet{openai2025misbehavior-cot-monitoring}.

In conclusion, our main contributions lie in the following three aspects:
\begin{packeditemize}
    \item We explore the cognitive habits exhibited by LRMs, which are consistent behavioral patterns that emerge independently of specific tasks.
    \item We introduce \ourbench, a principled benchmark grounded in the established cognitive habit framework developed for humans.
    \item We conduct a comprehensive analysis of 16 well-known LLMs, demonstrating that LRMs exhibit distinct cognitive habits. Furthermore, we extend our evaluation to safety-related contexts, revealing that certain cognitive habits are strongly associated with generating harmful responses.
\end{packeditemize}

\begin{figure*}[t]
    \centering
    \includegraphics[width=1\linewidth]{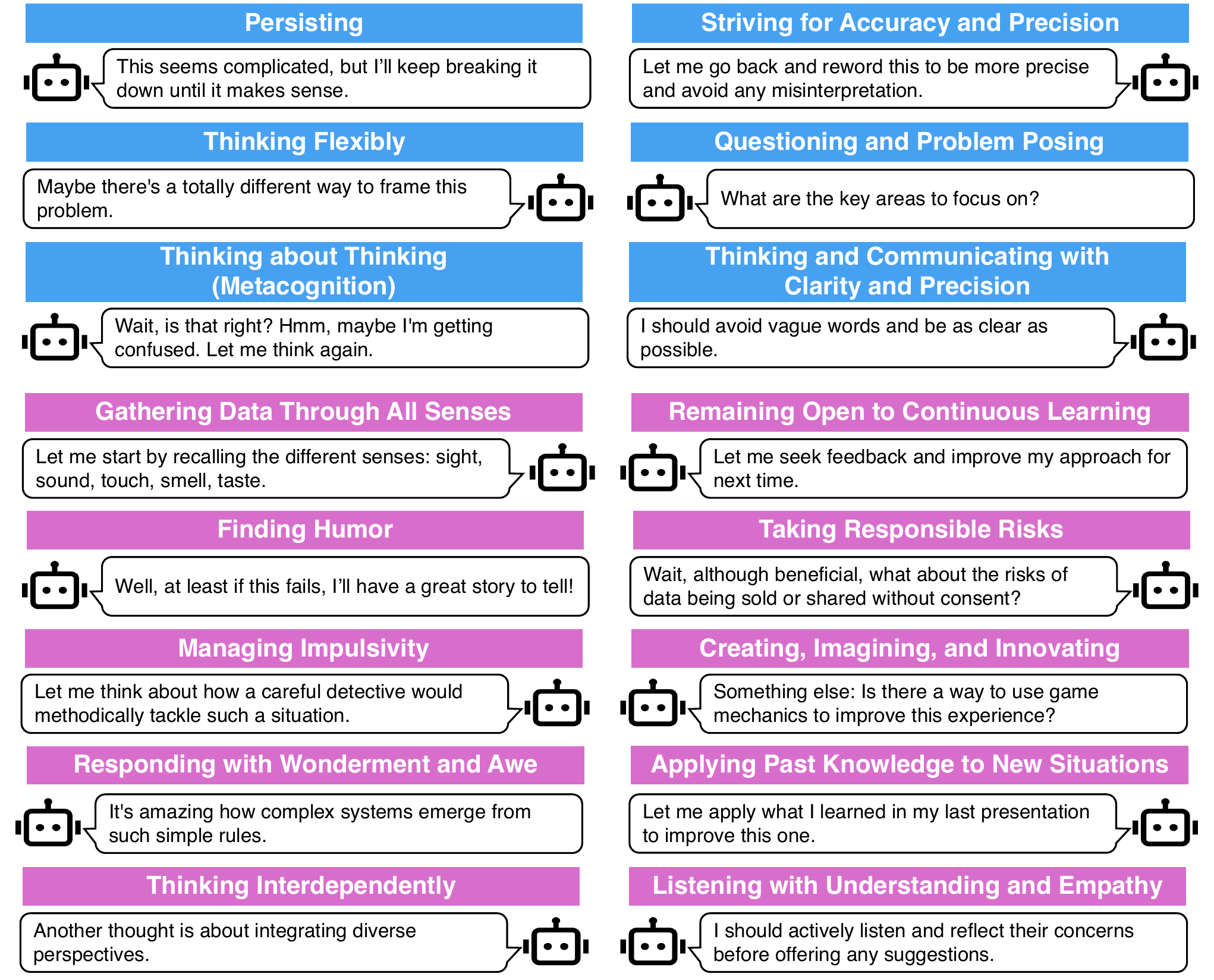}
    \caption{\textbf{The \textit{Habits of Mind} framework and the corresponding examples of meta-thinking statements}. Habits are evaluated via math problems (blue) and LLM-generated tasks (purple).
    }
    \label{fig:habit-of-mind-intro}
\end{figure*}

\section{Background}

\noindent \textbf{Large Reasoning Models (LRMs)}.
\citet{wei2022cot} pioneer eliciting reasoning Chain of Thoughts (CoTs) from auto-regressive LLMs~\citep{radford2018gpt1}, which has demonstrated effectiveness in boosting LLM performance on reasoning tasks~\citep{kojima2022zero-shot-cot,shi2023language-llm-are-multilingual-reasoners}.
Subsequent works~\citep{zelikman2022star,yu2024metamath-rationale-cot-fine-tuning-ref-1} focus on crafting high-quality CoT data to enhance LLMs' reasoning abilities via supervised learning.
However, obtaining supervised CoT data in a scalable fashion remains expensive due to its reliance on substantial human expertise.
Test-time compute scaling~\citep{snell2024scaling-test-time-compute-optimality,guan2025rstar-math-rationale-with-mcts-and-prm} offers another path by iteratively and reflectively guiding LLMs through the solution space, further improving reasoning performance.
Narrowly speaking, LRMs represent a distinct trajectory: resorting to reinforcement learning from verifiable rewards to teach LLMs about effective and long reasoning CoTs.
LRMs are inherently capable of generating an explicit reasoning CoT before generating a final response.
That said, LRMs decide their exploration of the solution space via CoT generation by themselves.
Representative LRMs include OpenAI's \llm{o1}~\citep{jaech2024openai-o1-system-card}, the open-source \llm{DeepSeek-R1}~\citep{guo2025deepseek-r1}, and Anthropic's \llm{Claude-3.7-sonnet}~\citep{claude-3-7-sonnet}.

\noindent \textbf{The \textit{Habits of Mind} framework}~\citep{costa2005habits-of-mind-book}.
This covers 16 advanced human cognitive behavior patterns designed to facilitate effective problem-solving. 
These habits support intelligent behavior in situations where solutions are not immediately evident, which is particularly relevant to LRMs when addressing reasoning-intensive tasks. 
We enumerate the 16 habits and examples of their meta-thinking statements in~\Cref{fig:habit-of-mind-intro}.
Notably, habits such as \habit{Thinking Flexibly} align closely with meta-cognitive expressions observed in \llm{DeepSeek-R1}, such as the phrase ``Another thought: maybe there's a way to reallocate existing resources rather than bringing in new ones.'' 
This connection motivates our systematic investigation into the extent to which LRMs exhibit the habits in the \textit{Habits of Mind} framework.

\section{Measuring Cognitive Habits of Large Reasoning Models}

Our assessment of the cognitive habits of LRMs consists of three stages, as illustrated in~\Cref{fig:pipeline}:
(1) \textbf{Task Construction}: We design tailored tasks intended to elicit thought-intensive reasoning from LRMs;
(2) \textbf{CoT Observation}: The LRM responds to each task prompt, enabling us to capture the intermediate reasoning underlying its problem-solving process;
(3) \textbf{Habit Extraction}: We apply an LLM-based automated evaluation to identify the presence of target cognitive habits within the CoTs.

\begin{center}
\begin{promptboxc}[prompt:applying-past-knowledge-guideline]{
Guideline for the \habit{Applying Past Knowledge to New Situations} habit}
{\fontsize{10}{9}\selectfont
You are required to generate tasks that can effectively differentiate between candidates' cognitive habits.\\
\\
In this case, the habit is \habit{Applying Past Knowledge to New Situations}, which involves:\\
- Drawing relevant insights, strategies, or principles from previous experiences or learning.\\
- Adapting familiar solutions to novel problems or unfamiliar domains.\\
- Recognizing analogies, patterns, or connections between past and present contexts.\\
- Using what is known while being open to new constraints or nuances.\\
\\
Your goal is to generate 25 task instructions that \textbf{naturally elicit this habit}, without revealing that knowledge transfer is being assessed.\\
\\
Guidelines:\\
- Each task instruction should follow the format: \textbf{[Previous Experience] [Task Background/Context] [Task Instruction]}\\
- The candidate should be placed in a \textbf{functional or goal-directed role} (e.g., manager, consultant, analyst, teacher) in a setting that presents a new challenge or problem to solve.\\
- You may \textbf{briefly state} that the candidate has prior experience in a different domain, but this should appear only as background — \textbf{do not suggest it is relevant or needed} for the current task.\\
- The task itself must focus entirely on the \textbf{new context}, with \textbf{no mention of prior experience in the instruction itself}.\\
- Do \textbf{not} use any phrasing that hints at knowledge transfer (e.g., “apply what you’ve learned,” “adapt previous strategies,” “draw from your experience,” etc.).\\
- The instruction should direct the candidate to perform a concrete action (e.g., create a plan, design a proposal, recommend steps), entirely within the new domain.\\
- The presence or absence of the habit should emerge from whether the candidate \textit{independently} uses past knowledge in their solution.
}
\end{promptboxc}
\end{center}

\subsection{Design Principles}
Our testing of the LRM's cognitive habits takes the following factors into account:
\begin{packeditemize}
    \item \textbf{Habit Specificity}: Cognitive habits may only arise under specific circumstances. This necessitates task designs that situate LRMs in conducive and suitable scenarios.
    \item \textbf{Spontaneity}: LLMs have a known tendency toward sycophancy~\citep{sharma2024towards-anthropic-understanding-sycophancy}. Explicit mention of cognitive habits in task instructions may bias the model’s responses. Therefore, we refrain from incorporating habit-oriented instructions and instead objectively observe the emergence of cognitive habits in the model's CoT.
    \item \textbf{Real-World Utility}: Although positive cognitive habits are widely desirable, they are of limited value if exhibited by LRMs only in artificial scenarios. By contrast, we require tasks to be grounded in realistic scenarios, ensuring that any observed cognitive habits are indicative of meaningful progress in real-world problem-solving.
    \item \textbf{Comprehensiveness}: We do require large-scale testing of the cognitive habits across diverse settings. This also challenges the automation level of the evaluation framework, as large-scale human evaluation would be expensive. 
    \item \textbf{Scalability}: The evaluation method must be inherently scalable, enabling continuous tracking and analysis of cognitive habit development as models evolve over time.
\end{packeditemize}

\begin{promptboxc}[prompt:habit-extraction]{Evidence-first habit extraction}
\fontsize{10}{10}\selectfont
You will be given a THINKING PASSAGE that describes a person's internal thinking while solving a problem. Your task is to determine whether a specific cognitive habit is \textbf{explicitly} demonstrated in the thought process.\\
\\
To make this judgment, you must first identify and extract a \textbf{verbatim statement} from the THINKING PASSAGE that \textbf{clearly and directly} supports the presence of the specified habit.\\
\\
\#\#\# Habit to evaluate:\\
\textcolor{blue}{\{habit\}}\\
\\
\#\#\# Examples of statements that would support this habit:\\
\textcolor{blue}{\{example\_meta\_thinking\_statements\}}\\
\\
Your response must strictly follow this JSON format:\\
\{\{\\
    "evidence": "Exact sentence from the THINKING PASSAGE that directly demonstrates the habit, or an empty string if no such sentence exists",\\
    "is\_reflected": true/false (true if the evidence sentence directly demonstrates the habit, false otherwise)\\
\}\}\\
\\
\#\#\# Instructions (read carefully):\\
- Your answer must be based \textbf{only on explicit statements} from the THINKING PASSAGE. Do not infer, interpret, or assume anything beyond what is written.\\
- Do \textbf{not} reword the evidence. Use \textbf{exact text only} in the "evidence" field except empty string.\\
- If multiple statements support the habit, extract \textbf{only the first full sentence (in order of appearance)} that clearly and directly demonstrates it — even if only part of that sentence reflects the habit.\\
- If \textbf{no statement demonstrates} the habit, leave the "evidence" field empty and set "is\_reflected" to `false`.\\
- The presence of other habits in the same sentence is acceptable — your task is to judge \textbf{only whether the specified habit is present}.\\
- Responses that rely on vague or implicit reasoning will be considered invalid.
\end{promptboxc}

\subsection{Principled Testing of Cognitive Habits}

\noindent \textbf{Task Construction}.
We construct habit-tailored tasks using a hybrid approach that emphasizes \textit{Habit Specificity}.
For cognitive habits related to thinking, we recognize math problems as an ideal testbed.
Most such tasks are instantiated using the \dataset{MATH-500}~\citep{hendrycks2measuring-math-500}, except for the \habit{Persisting} habit, which we represent with more challenging problems drawn from \dataset{AIME}.
Given the demonstrated strengths of LLMs in instruction generation~\citep{wang2023self-instruct-model-self-training-1,yu2024metamath-rationale-cot-fine-tuning-ref-1}, we leverage advanced LLMs (\eg, \llm{GPT-4.1}) to generate tasks for the remaining habits.
To guide this process, two authors independently design detailed guidelines for each habit, which is a one-time effort.
An example of such a guideline, corresponding to the \habit{Applying Past Knowledge to New Situations} habit, is presented in~\Cref{prompt:applying-past-knowledge-guideline}.
To ensure alignment with the principles of \textit{Spontaneity} and \textit{Real-World Utility}, we explicitly incorporate constraints into the task prompts.
While the above curation methodology enjoys the benefits of \textit{Scalability}, we complement it with manual quality verification to ensure reliability.
We construct a diverse set of 25 tasks for each of the 16 cognitive habits, ensuring alignment with the principle of \textit{Comprehensiveness}.
We refer to this benchmark as \ourbench, which lays the foundation for our empirical study.

\noindent \textbf{CoT Observation}.
CoT has demonstrated effectiveness in boosting LLMs' abilities across a variety of tasks~\citep{wei2022cot, kojima2022zero-shot-cot,wang2024CoTdecoding}.
Advancing in this direction, LRMs are trained to generate responses through an explicit and intrinsic CoT reasoning process.
Compared to final answers, the generated CoTs offer deeper insights into the rationale underlying LRMs’ decision-making and problem-solving.
This transparency facilitates behavioral monitoring of LRMs, such as identifying potential safety concerns~\citep{openai2025misbehavior-cot-monitoring, chen2025reasoning-anthropic-unfaithful-cot-for-monitoring}.
Echoing the \textit{Spontaneity} principle, we query LRMs with habit-specific tasks without any additional intervention.
In this way, we obtain the CoT behind LRM's treatment of each given task, which can closely reflect the inherent reaction of LRMs when exposed to tasks mirroring real-world scenarios.
We further investigate the differences between the intrinsic CoTs of LRMs and the elicited CoTs produced by non-reasoning LLMs.
This comparison sheds light on why LRMs often substantially outperform their non-reasoning counterparts.

\begin{promptbox}[prompt:thinking-with-chat-models]{Responding after thinking}
\fontsize{10}{10}\selectfont
A conversation between User and Assistant. The user asks a question, and the Assistant solves it. The assistant first thinks about the reasoning process in the mind and then provides the user with the answer. The reasoning process and answer are enclosed within <think> </think> and <answer> </answer> tags, respectively, i.e., <think> reasoning process here </think> <answer> answer here </answer>.
\end{promptbox}

\noindent \textbf{Habit Extraction}.
To automate the identification of cognitive habits in CoTs, we leverage prompted LLMs as annotators.
We formulate this task as a binary classification problem: given a CoT and a specified target cognitive habit, the model determines whether the habit is present.
Our close observation of CoTs reveals that such habits are often manifested through key meta-cognitive statements, as exemplified in~\Cref{fig:habit-of-mind-intro}.
To mitigate hallucinations~\citep{zhang2023siren-song-survey-hallucination-llms} and enhance judgment reliability, we adopt an evidence-based annotation paradigm, using~\Cref{prompt:habit-extraction}.
Specifically, the annotator model is prompted to first extract supporting meta-cognitive statements before making a final determination regarding the habit's presence. 
Empirically, this evidence-first approach not only improves the performance of the annotator but also provides greater accountability.
One benefit of the design is that even a weak annotator model can supervise the cognitive habits of stronger LRMs, facilitating scalable oversight~\citep{bills2023language-llm-can-explain-neurons-in-llm}.

\begin{figure*}[!t]
    \centering
    \begin{subfigure}{0.86\textwidth}
        \includegraphics[width=\linewidth]{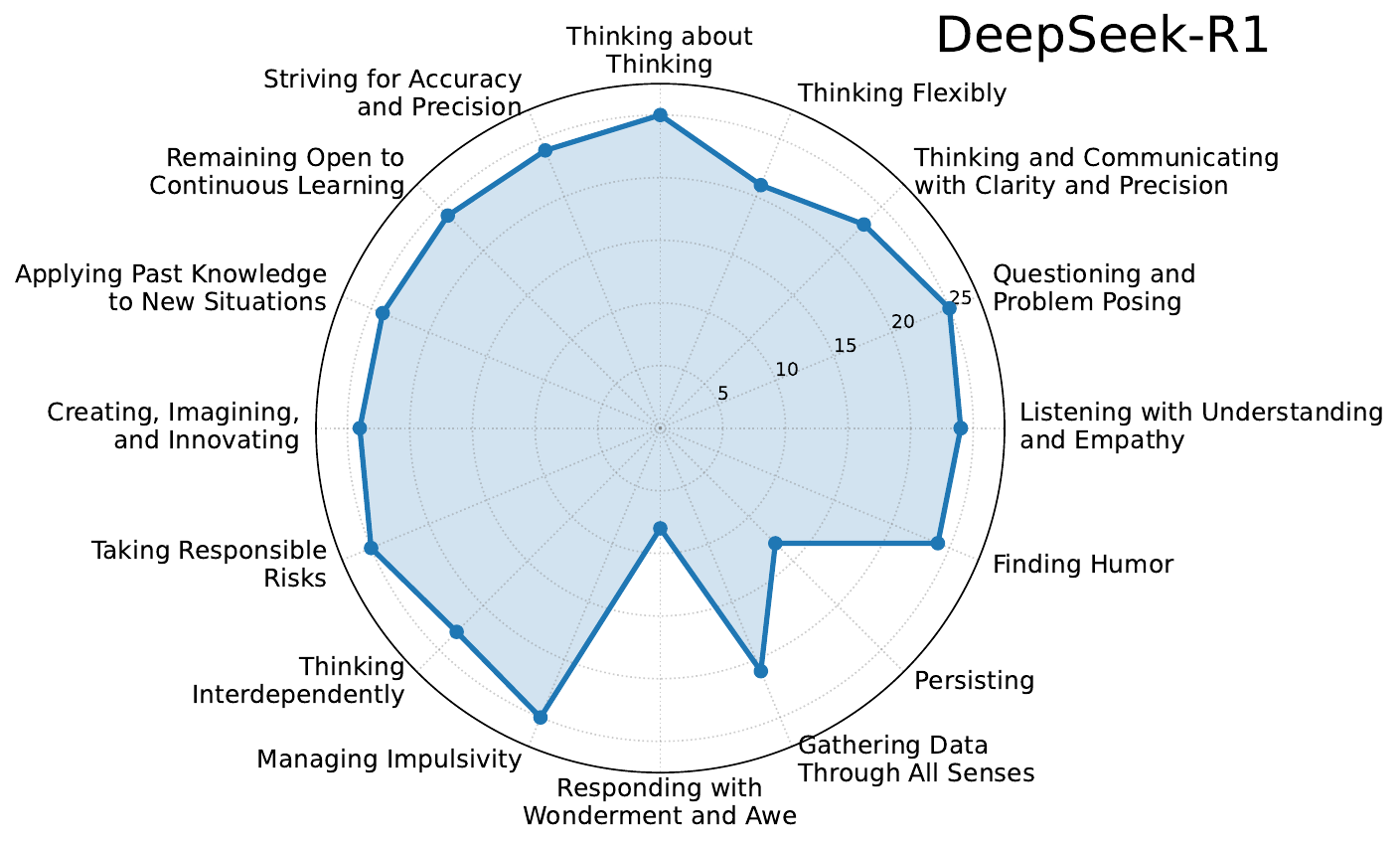}
    \end{subfigure}
    \vspace{0.5em}

    \begin{subfigure}{0.19\textwidth}
        \includegraphics[width=\linewidth]{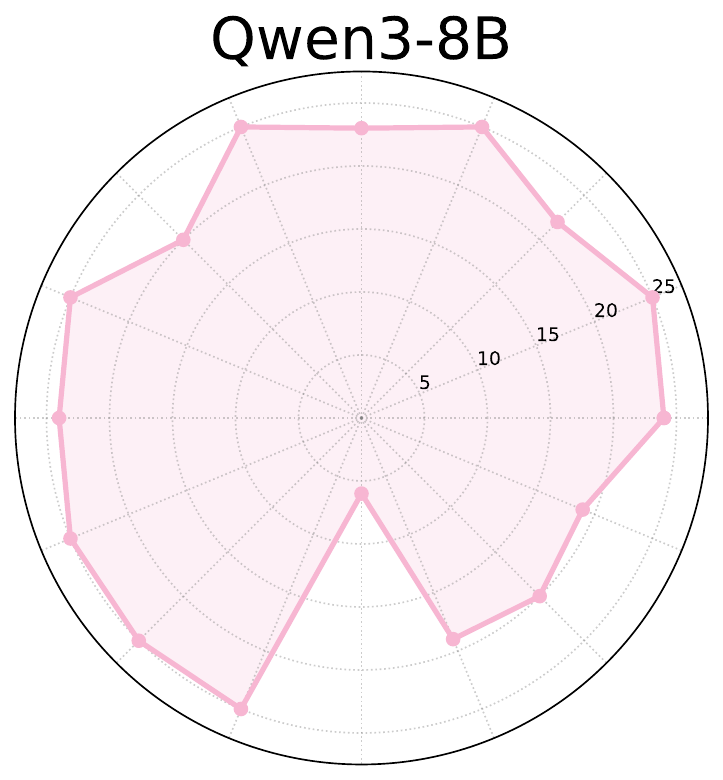}
    \end{subfigure}
    \begin{subfigure}{0.19\textwidth}
        \includegraphics[width=\linewidth]{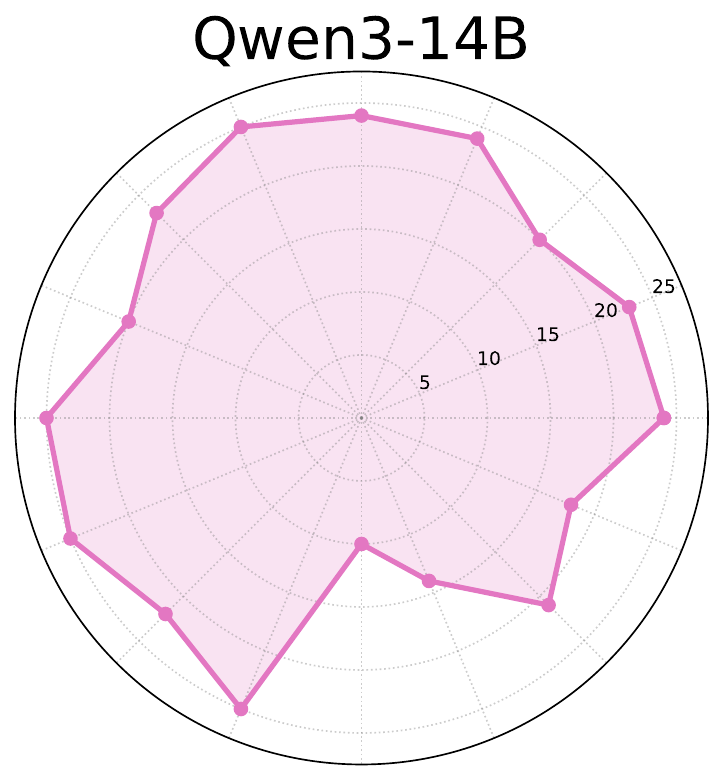}
    \end{subfigure}
    \begin{subfigure}{0.19\textwidth}
        \includegraphics[width=\linewidth]{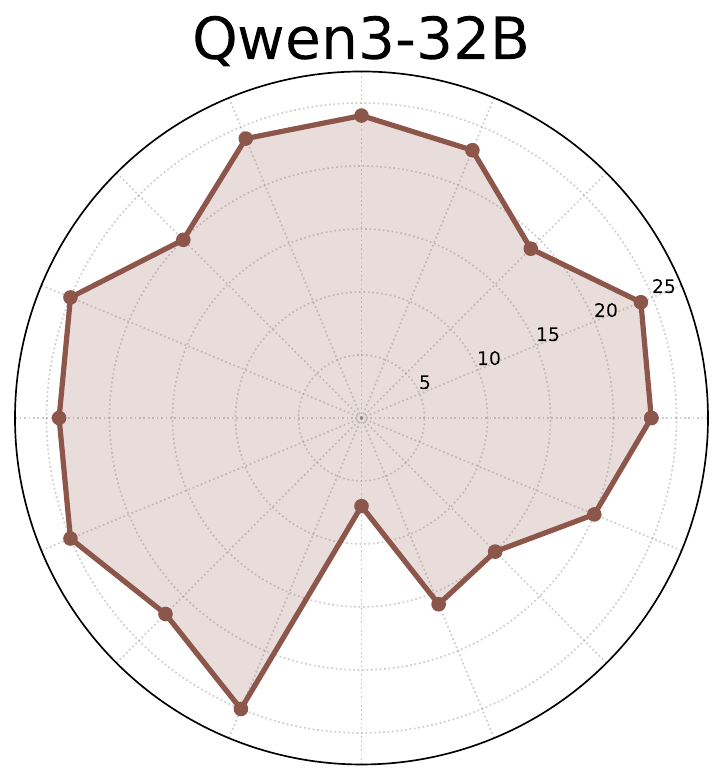}
    \end{subfigure}
    \begin{subfigure}{0.19\textwidth}
        \includegraphics[width=\linewidth]{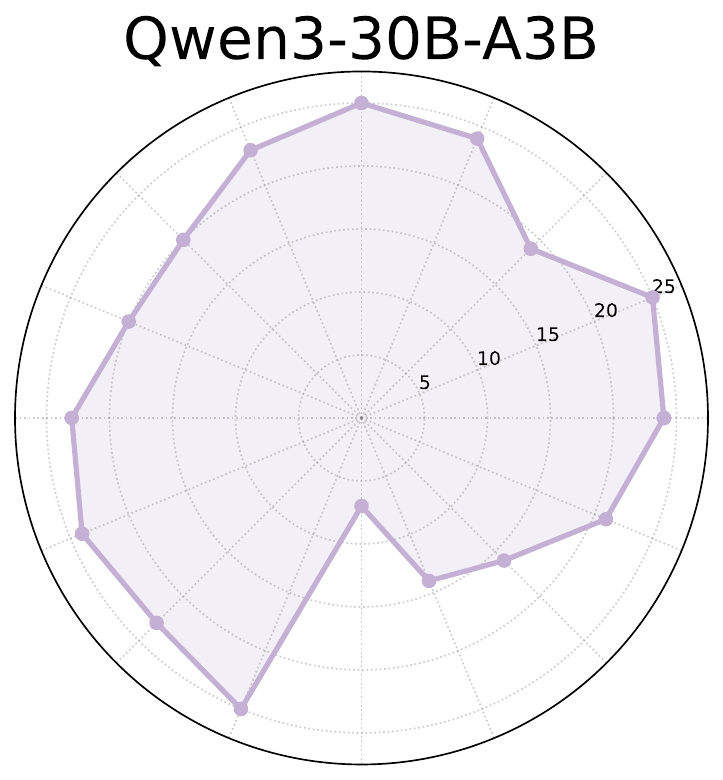}
    \end{subfigure}
    \begin{subfigure}{0.19\textwidth}
        \includegraphics[width=\linewidth]{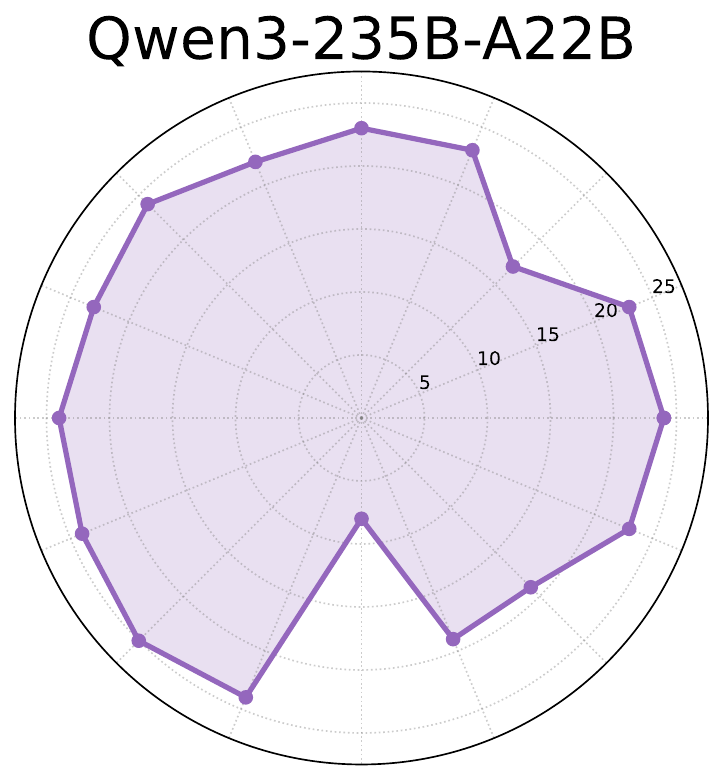}
    \end{subfigure}

    \vspace{0.5em}
    \begin{subfigure}{0.19\textwidth}
        \includegraphics[width=\linewidth]{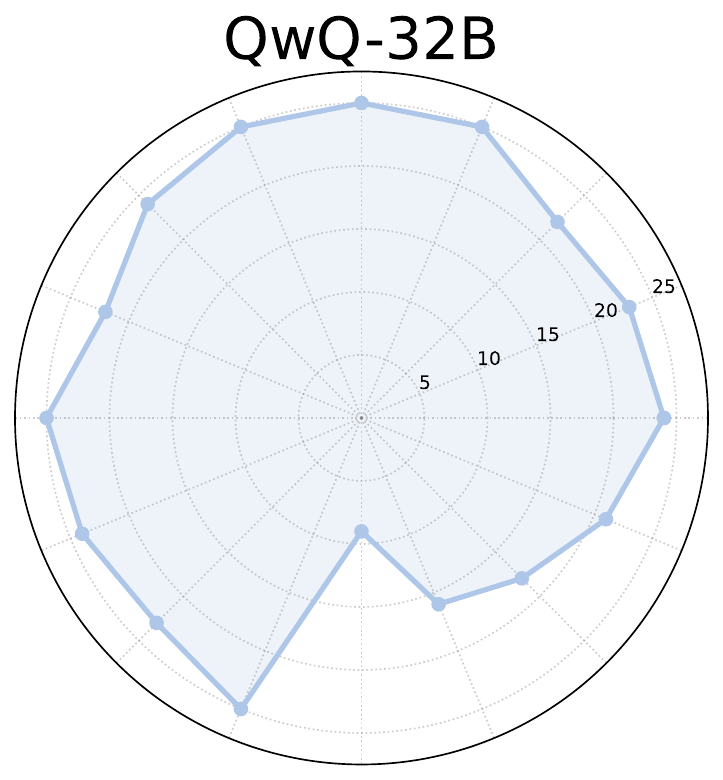}
    \end{subfigure}
    \begin{subfigure}{0.19\textwidth}
        \includegraphics[width=\linewidth]{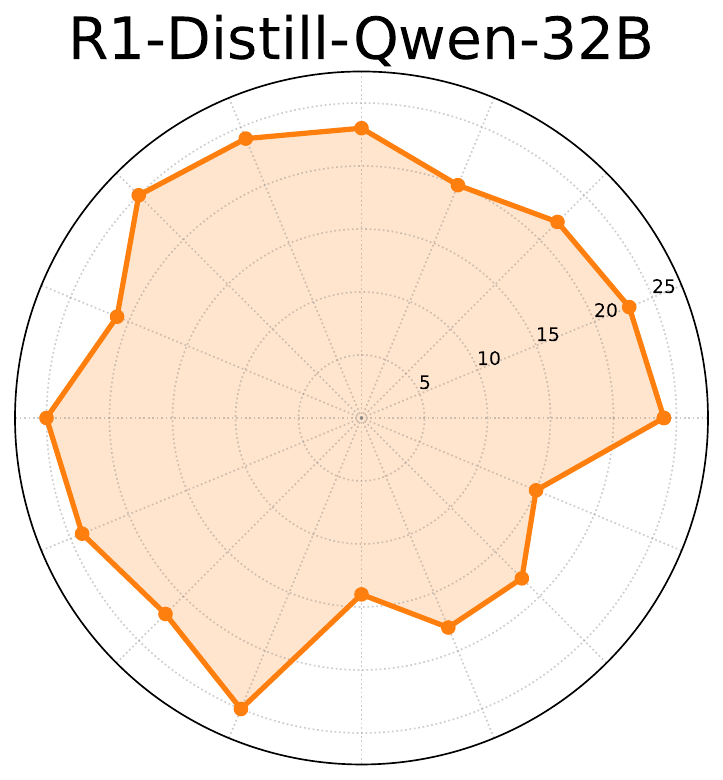}
    \end{subfigure}
    \begin{subfigure}{0.19\textwidth}
        \includegraphics[width=\linewidth]{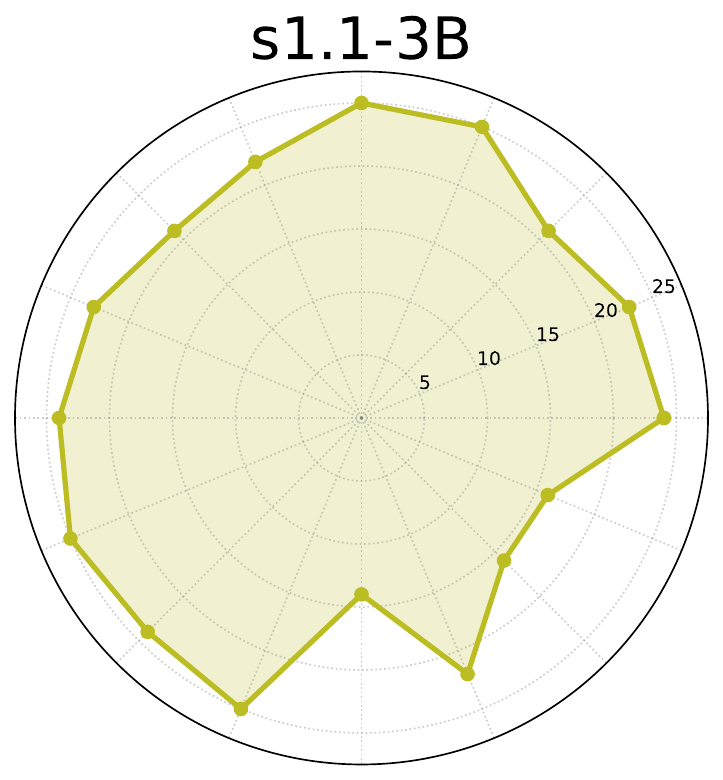}
    \end{subfigure}
    \begin{subfigure}{0.19\textwidth}
        \includegraphics[width=\linewidth]{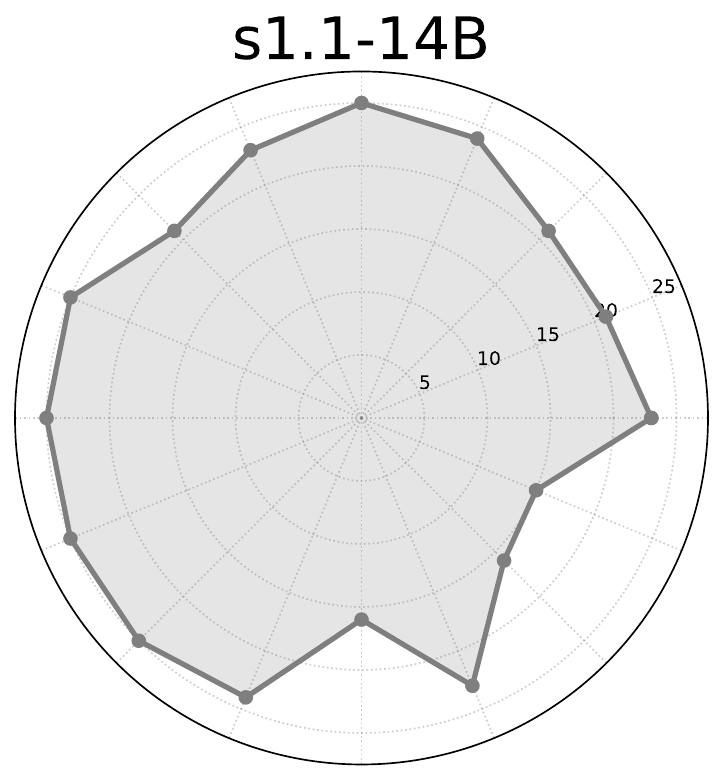}
    \end{subfigure}
    \begin{subfigure}{0.19\textwidth}
        \includegraphics[width=\linewidth]{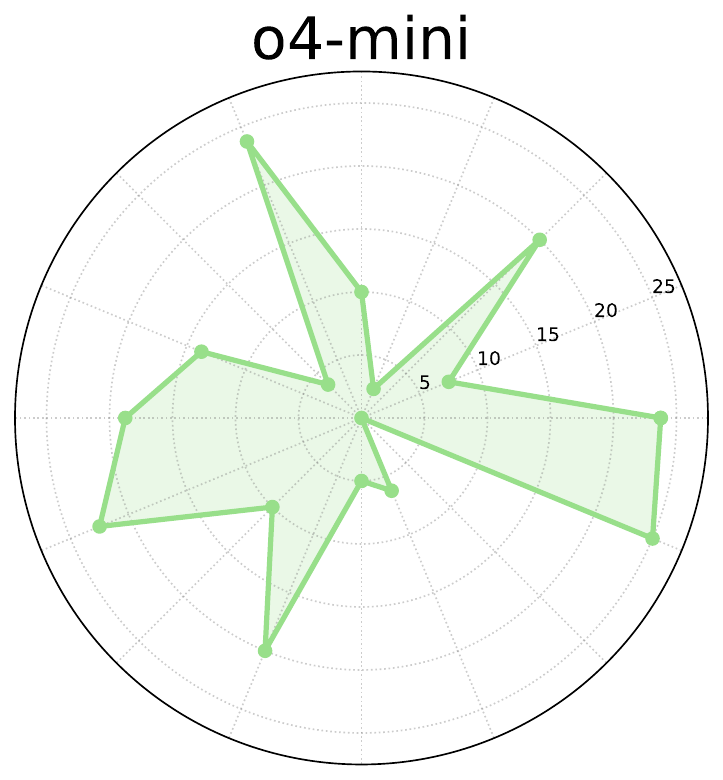}
    \end{subfigure}

    \vspace{0.5em}
    \begin{subfigure}{0.19\textwidth}
        \includegraphics[width=\linewidth]{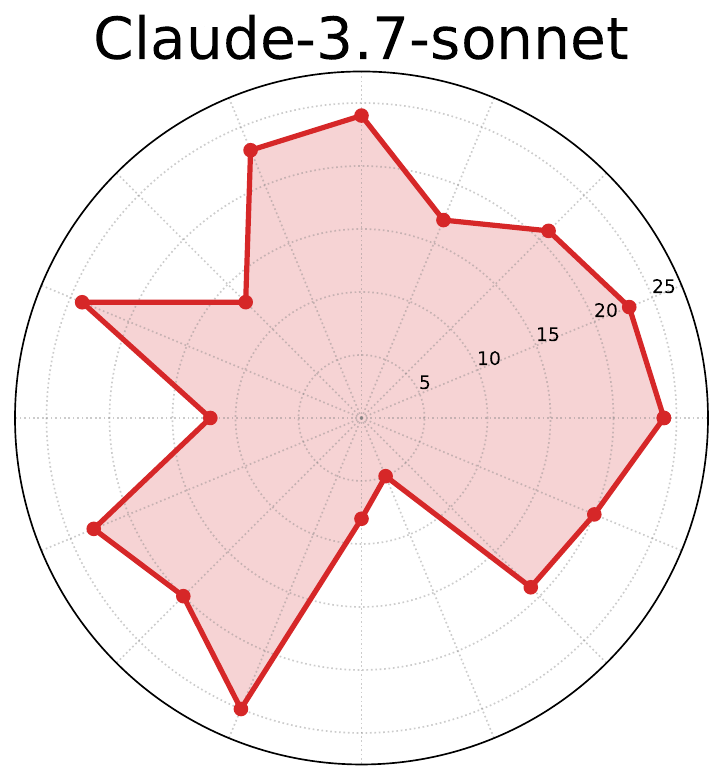}
    \end{subfigure}
    \begin{subfigure}{0.19\textwidth}
        \includegraphics[width=\linewidth]{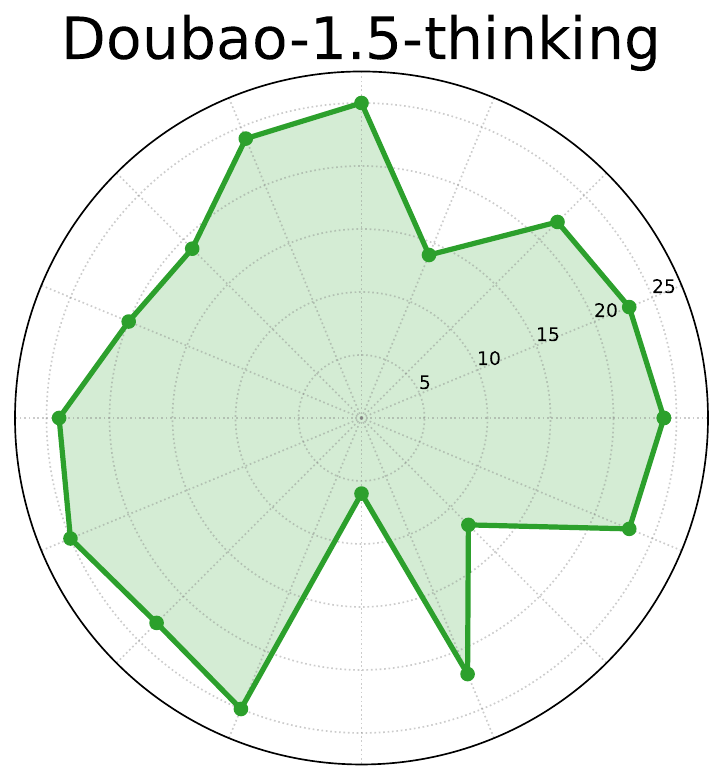}
    \end{subfigure}
    \begin{subfigure}{0.19\textwidth}
        \includegraphics[width=\linewidth]{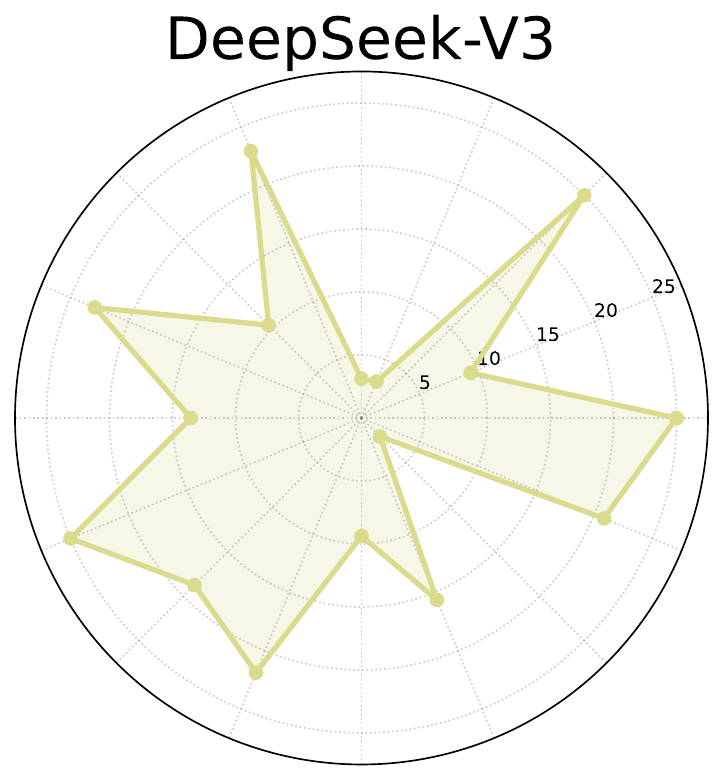}
    \end{subfigure}
    \begin{subfigure}{0.19\textwidth}
        \includegraphics[width=\linewidth]{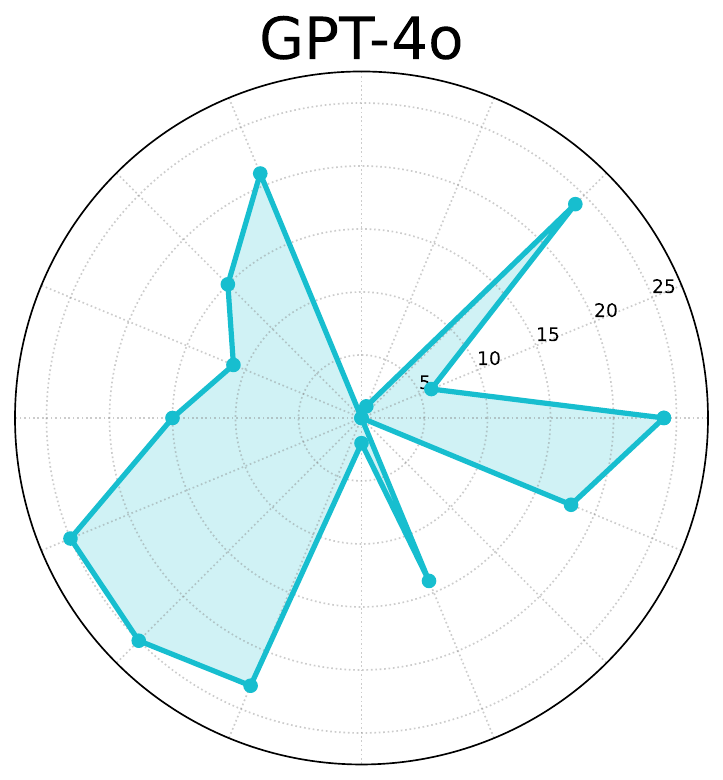}
    \end{subfigure}
    \begin{subfigure}{0.19\textwidth}
        \includegraphics[width=\linewidth]{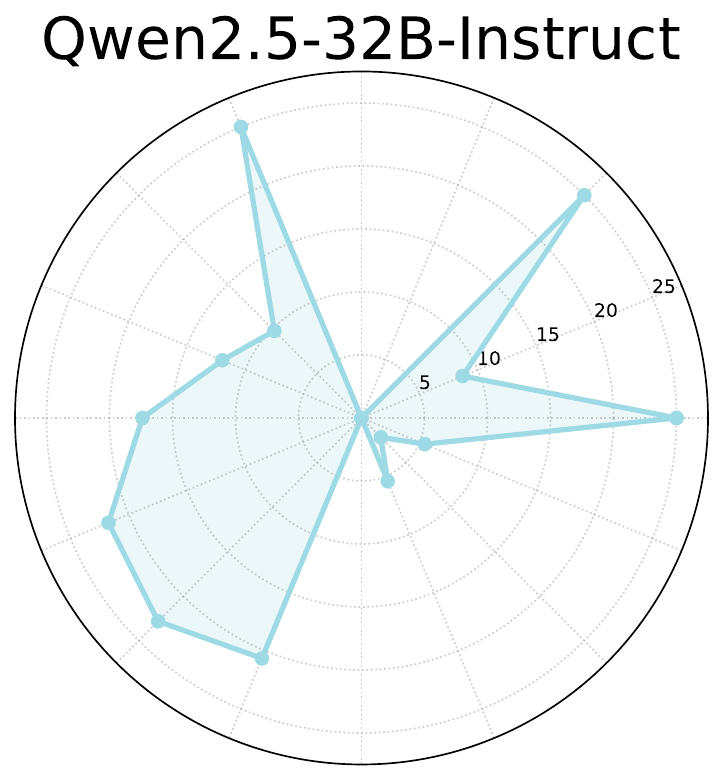}
    \end{subfigure}

    \caption{\textbf{Measurement results of the 16 LRMs' cognitive habits on \ourbench.} All other LRMs, with habit names omitted for brevity, follow the same habit display ordering as \llm{DeepSeek-R1}.}
    \label{fig:individual_radars}
\end{figure*}

\subsection{Experimental Setup}

\noindent \textbf{Evaluated Models}.
In this work, we cover a total of 16 well-recognized LLMs from three representative types of model candidates:
\begin{packeditemize}
    \item \textbf{Ten open-source LRMs} include \llm{DeepSeek-R1}~\cite{guo2025deepseek-r1}, \llm{DeepSeek-R1-Distill-Qwen-32B}~\citep{guo2025deepseek-r1}, \llm{Qwen-3} models (8B/14B/32B/30B-A3B/235B-A22B)~\citep{qwen2025qwen3technicalreport}, \llm{QwQ-32B}~\citep{qwq32b}, and \llm{s1.1-3B/14B}~\citep{muennighoff2025s1}.
    \item \textbf{Three closed-source LRMs} include \llm{o4-mini}~\citep{o3-and-o4-mini}, \llm{Claude-3.7-sonnet}~\citep{claude-3-7-sonnet}, and \llm{Doubao-1.5-thinking-pro}~\citep{seed2025seed-thinking-v1-5}.
    As vendors like OpenAI only return the summary of CoT contents\footnote{\url{https://platform.openai.com/docs/guides/reasoning\#reasoning-summaries}}, our testing on them only serves as a lower bound of exhibited cognitive habits.
    \item  \textbf{Three CoT-requested LLMs} consist of \llm{DeepSeek-V3}~\citep{liu2024deepseek-v3}, \llm{GPT-4o}~\citep{hurst2024gpt-4o-system-card}, and \llm{Qwen-2.5-32B-Instruct}~\citep{qwen2024qwen25technicalreport}. To align with LRMs, we prompt them to explicitly produce a prior-generation CoT, using~\Cref{prompt:thinking-with-chat-models}.
\end{packeditemize}

\begin{figure*}[!t]

    \begin{subfigure}{0.5\textwidth}
        \includegraphics[width=\linewidth]{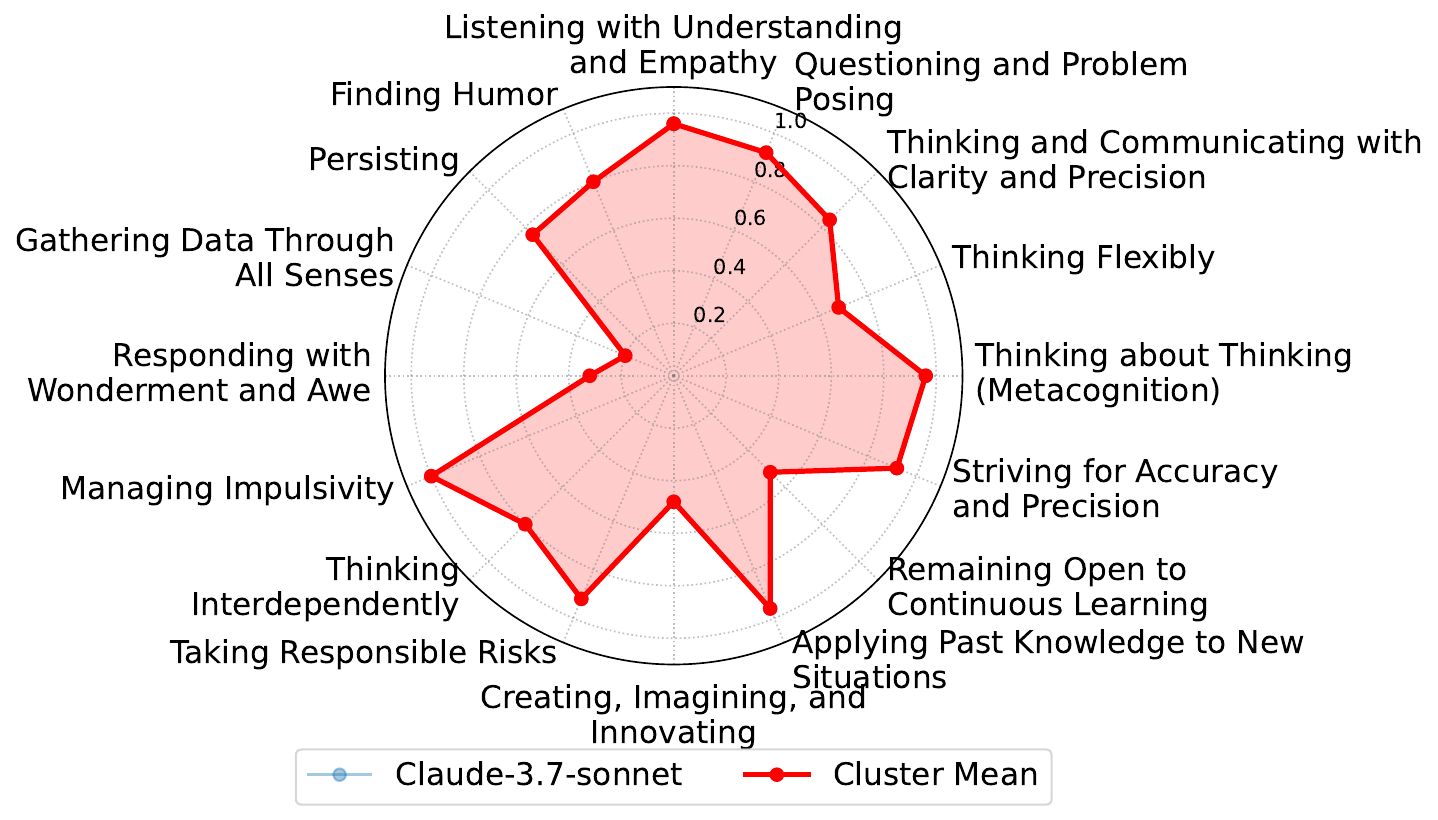}
    \end{subfigure}
    \begin{subfigure}{0.5\textwidth}
        \includegraphics[width=\linewidth]{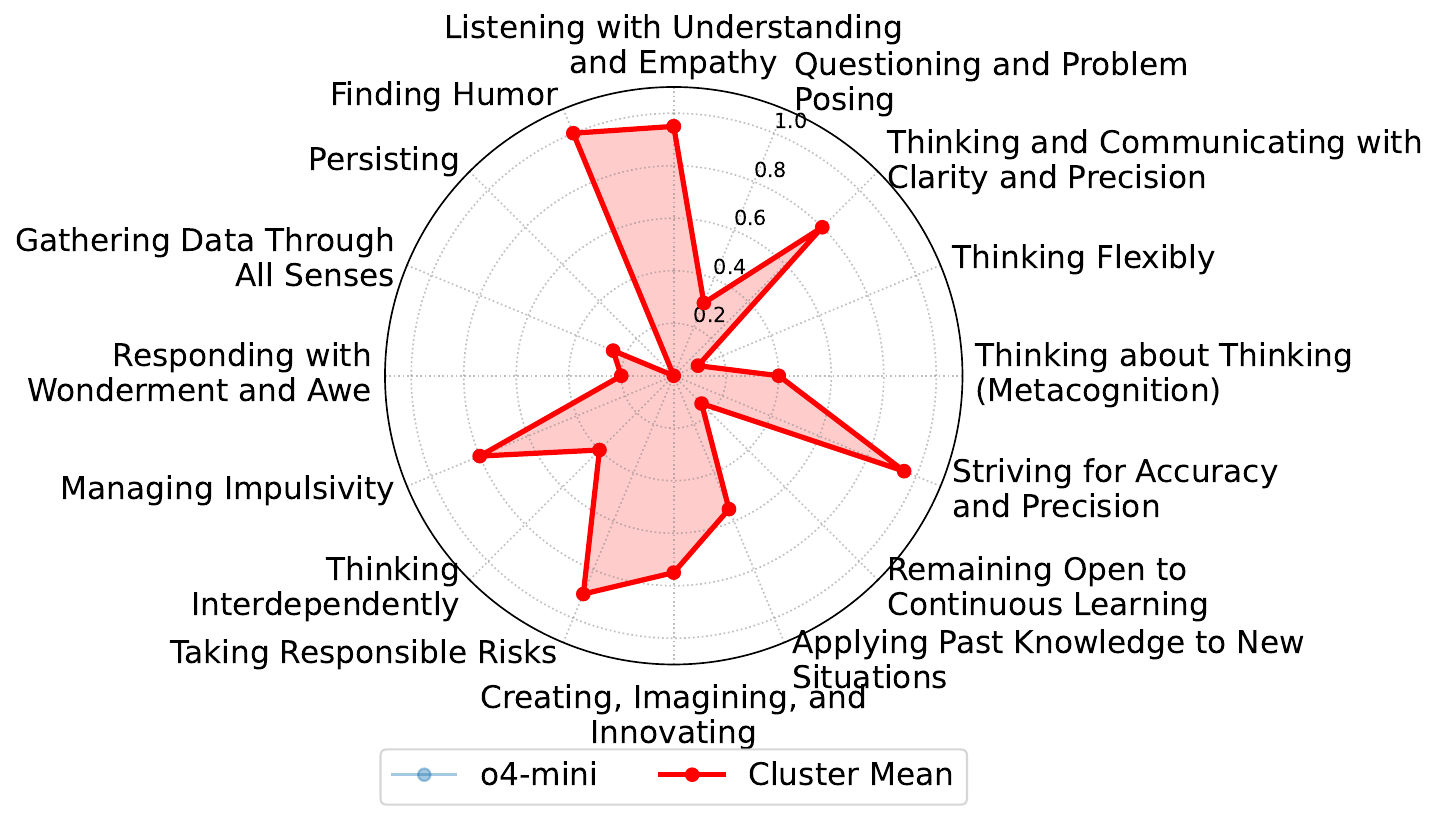}
    \end{subfigure}

    \vspace{0.5em}
    \begin{subfigure}{0.5\textwidth}
        \includegraphics[width=\linewidth]{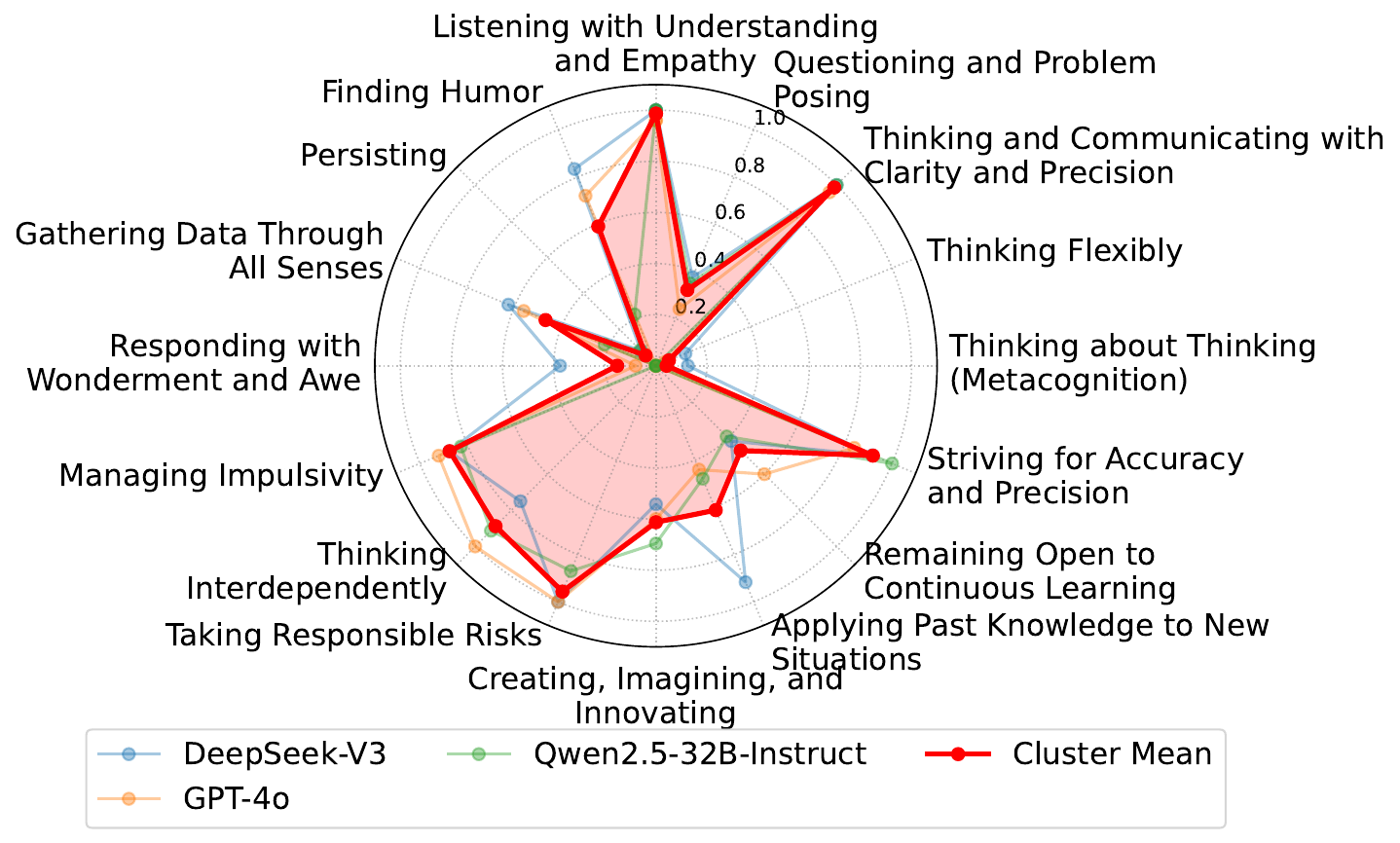}
        \vspace{0.2em}
    \end{subfigure}
    \begin{subfigure}{0.5\textwidth}
        \includegraphics[width=\linewidth]{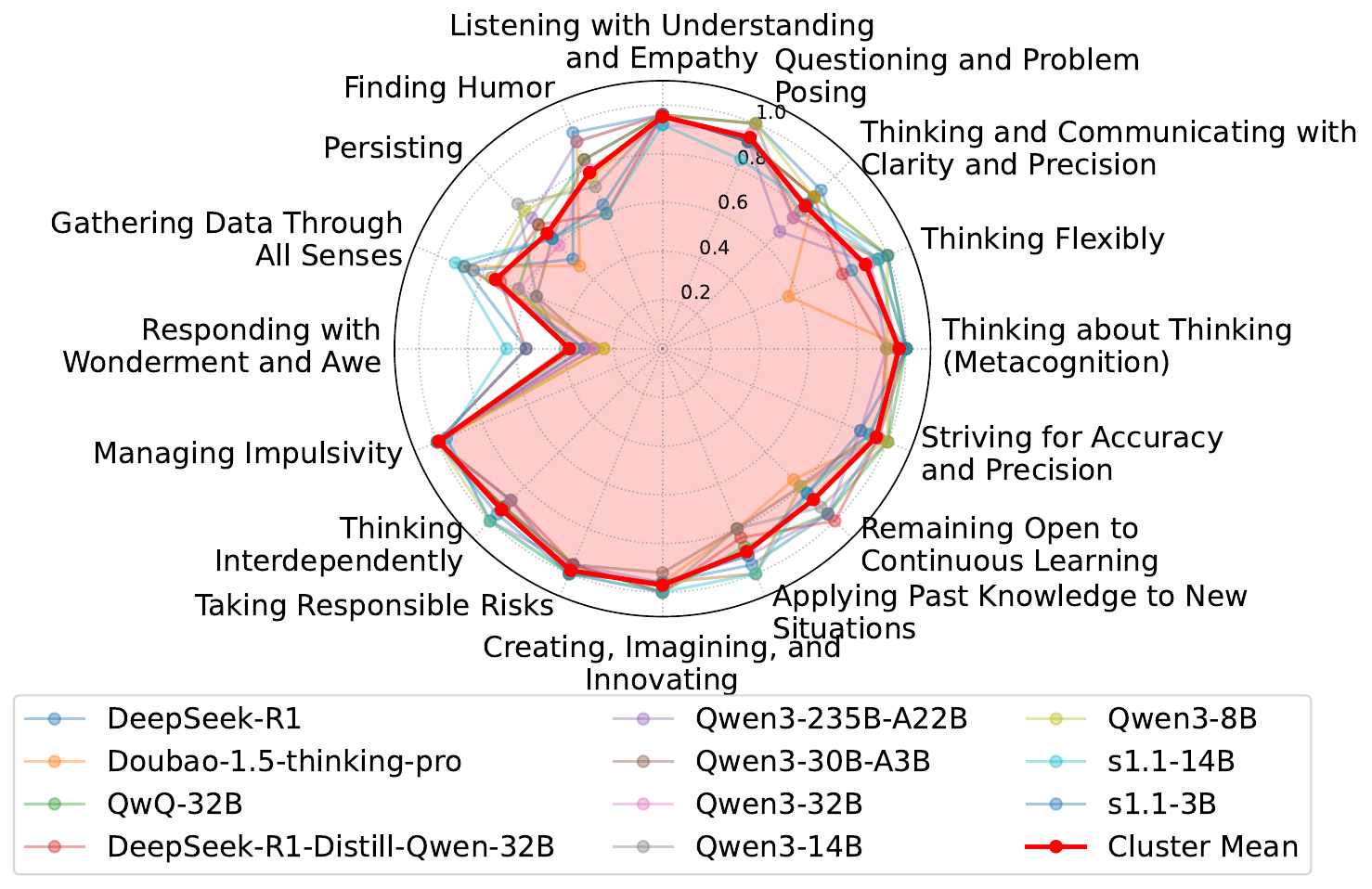}
    \end{subfigure}
    \caption{\textbf{Measurement results of the 16 LRMs' cognitive habits on \ourbench.} All other LRMs, with habit names omitted for brevity, follow the same habit display ordering as \llm{DeepSeek-R1}.}
    \label{fig:cluster}
\end{figure*}
\noindent \textbf{CoT Generation}.
For closed-source LRMs, we query their official APIs and obtain the CoTs or provided CoT summaries.
For open-source LLMs, we locally deploy the models, using vLLM~\citep{kwon2023vllm} for accelerated inference.
We follow the official chat templates of the models.
Following suggested practices, we set the temperature to 0.6 and the top\_p to 0.95.

\noindent \textbf{Habit Extraction}. We employ \llm{GPT-4.1-mini} as the annotation model, which enjoys the benefits of both effectiveness and cost efficiency.
We provide the definition and examples of corresponding meta-thinking statements for calibration.
The temperature is set to 0.0 and top\_p to 0.95.

\subsection{Measurement Results}

\noindent \textbf{LRMs exhibit diverse cognitive habits, persistent across tasks.}
We visualize the measurement results in~\Cref{fig:individual_radars}. 
Taking \llm{DeepSeek-R1} as an example, this model demonstrates certain positive habits when faced with tasks that can benefit from them. 
The CoT analysis reveals that LLMs' problem-solving processes go beyond merely inferring user intent, encompassing behaviors guided by ingrained cognitive habits. 
What's more, the cognitive behaviors persist across different tasks, establishing cognitive habits that we study in this work.
Notable differences are observed in the extent to which distinct cognitive habits are possessed.
As LRMs are typically trained to excel in reasoning-intensive tasks, it is within the expectation that they tend to exhibit habits such as \habit{Striving for Accuracy and Precision} and \habit{Thinking about Thinking}, both of which are highly relevant for exploring solution spaces~\citep{snell2024scaling-test-time-compute-optimality}. 
Surprisingly, even the habits that are not directly associated with reasoning are observed in LRMs as well. 
For instance, the habits \habit{Applying Past Knowledge to New Situations} and \habit{Remaining Open to Continuous Learning} suggest adaptability and receptiveness to new information; 
\textit{Finding Humor} and \habit{Listening with Understanding and Empathy} indicate a capacity for affective and social sensitivity; 
\habit{Gathering Data Through All Senses} is essential for multi-modal foundation models~\citep{liu2024visual-llava,fei2025pathmultimodalgeneralistgenerallevel}. 
However, we also observe that models like \llm{DeepSeek-R1} are weak in \habit{Responding with Wonderment and Awe} (8 out of 25), reflecting a tendency to operate in a highly confident manner. 
This behavior may be associated with poor calibration~\citep{geng2024survey-llm-calibration}, highlighting a promising direction aimed at improving LLM alignment.

\begin{figure*}[tbp]
    \centering
    \includegraphics[width=1\linewidth]{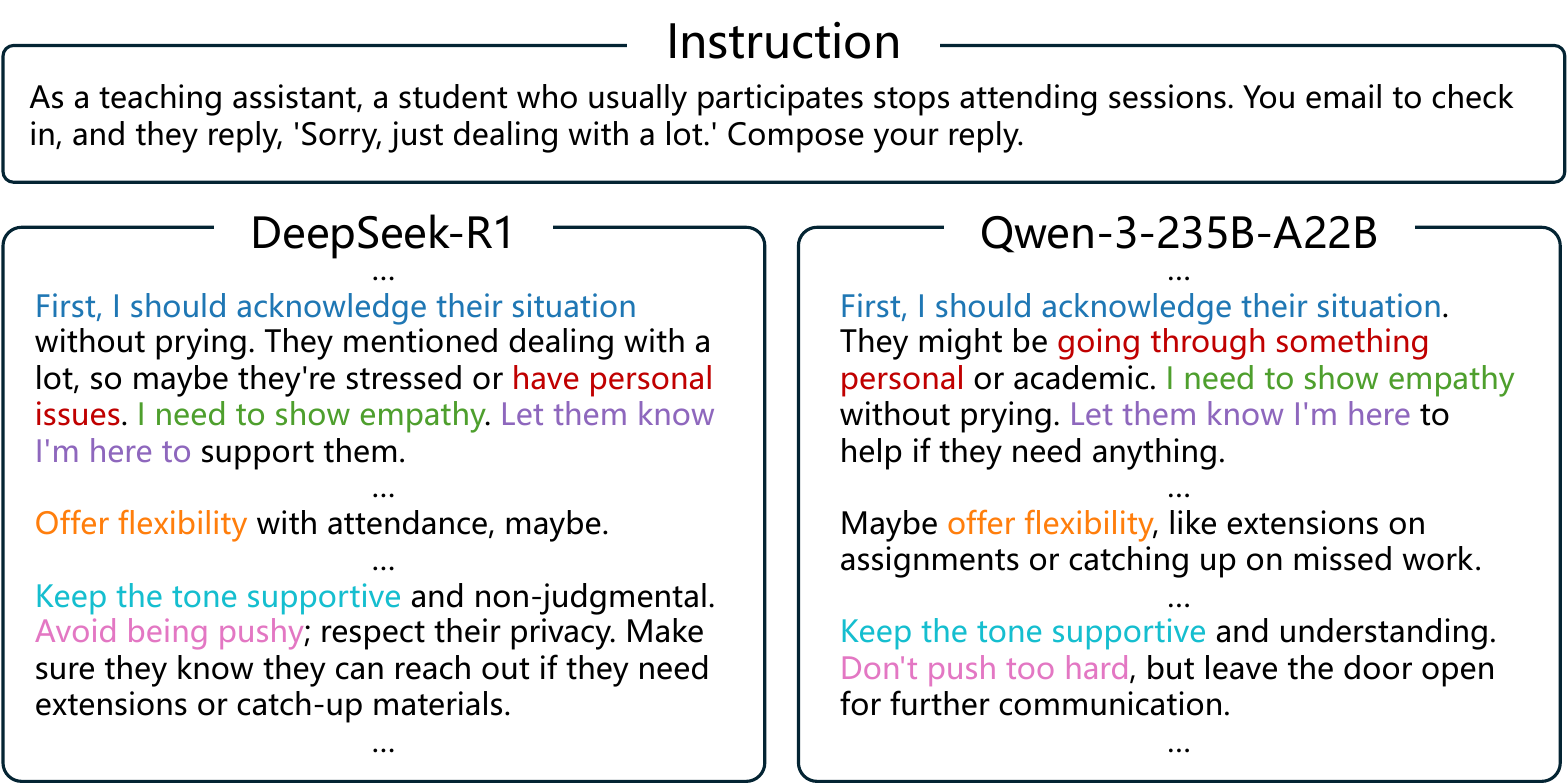}
    \caption{\textbf{Qualitative example of similarity between \llm{DeepSeek-R1} and \llm{Qwen-3-235B-A22B} in CoT trajectories and certain steps}. This task belongs to the \habit{Listening with Understanding and Empathy } habit.
    }
    \label{fig:qualitative-example-of-similar-cot}
\end{figure*}

We define cognitive habit profiles as the frequency with which cognitive habits are activated on \ourbench by a given LRM.
Another intriguing observation reveals notable similarities and differences across models.
To systematically analyze these similarities, we apply agglomerative clustering~\citep{murtagh2014ward-agglomerative-clustering} to the cognitive habit profiles of LRMs into 4 clusters.

\noindent \textbf{LRMs vs. Non-Reasoning Models in Cognitive Habits}.
Recall that we prompt non-reasoning LLMs to generate responses after thinking, resulting in superficial programmatic similarities to LRMs.
However, as illustrated in~\Cref{fig:cluster}, their cognitive habit profiles remain markedly distinct from those of LRMs, particularly in habits essential for reasoning-intensive tasks, such as \habit{Thinking about Thinking}.
Meanwhile, non-reasoning models do exhibit certain cognitive habits; for example, all three LLMs show a strong tendency toward the habit of \habit{Thinking and Communicating with Clarity and Precision}, consistent with findings from~\citet{wang2025thoughts-are-all-over-the-place-underthinking}.
Nonetheless, it is worth noting that an implicit limitation is their inherent inability to generate long CoTs, which undermines their performance on complex reasoning tasks.
While the empirical findings on \ourbench do not comprehensively capture all the advantages of LRMs over non-reasoning models, they offer compelling evidence for the benefits of reasoning RL in cultivating effective problem-solving habits in LRMs.

\noindent \textbf{Comparative Analyses of Cognitive Habits Across LRMs}.
The LRMs cluster into three groups, with closed-source models such as \llm{o4-mini} and \llm{Claude-3.7-sonnet} exhibiting clear dissimilarities from the others.
\llm{Claude-3.7-sonnet}, despite its ability to handle multimodal inputs~\citep{claude-3-7-sonnet}, demonstrates weak performance in the habit of \habit{Gathering Data Through All Senses}.
This suggests that although the model possesses strong multimodal capabilities, it may fail to proactively utilize all modalities in certain tasks, indicating a potential capacity-utility gap.

Due to OpenAI's restriction to summaries of reasoning CoTs, we can only estimate a lower bound on the cognitive habits exhibited by GPT models, as meta-thinking statements may be omitted from the CoT summaries.\footnote{However, this limitation does not prevent OpenAI's internal developers from evaluating the cognitive behaviors of GPT models using \ourbench.}
Even so, we observe that \llm{o4-mini} tends to produce contextually appropriate and user-aligned responses, consistent with OpenAI’s model specifications~\citep{openai-model-spec}.

Models within the same family, such as the \llm{Qwen-3} series with different parameter sizes, exhibit highly similar cognitive habit profiles.
This suggests that the formation of cognitive habits can be largely influenced by the underlying training algorithms and data.
Similarly, LLMs distilled from \llm{DeepSeek-R1}, including \llm{R1-Distill-Qwen-32B} and the \llm{s1.1} models, display closely aligned cognitive habit profiles.

A broader analysis of LRMs in~\Cref{fig:cluster} reveals an interesting phenomenon: models from different families can also exhibit similar cognitive habit profiles.
For instance, we observe a notable resemblance between the \llm{Qwen3} models and \llm{DeepSeek-R1}.
More intriguingly, a qualitative analysis of their CoTs reveals strikingly similar reasoning steps and highly analogous CoT trajectories on certain tasks, as exemplified in~\Cref{fig:qualitative-example-of-similar-cot}.
Inspired by these intra-family and post-distillation similarities, we cautiously hypothesize that the observed resemblance between \llm{Qwen3} models and \llm{DeepSeek-R1} may stem from \textbf{biases in training algorithms} or \textbf{unintentional data contamination} (see~\Cref{subsec:lrm-resemblance}).
We leave a detailed exploration of this phenomenon for our future work.

\section{Case Study: Safety}
We examine the cognitive habits exhibited by LRMs when responding to safety-related user queries. 
Specifically, we investigate whether LRMs tend to generate CoTs with certain identifiable cognitive habits when generating harmful versus harmless responses.

\newcommand{\hhratio}[2]{\vspace{0.5em} \shortstack[l]{Harmful: \textcolor{red}{#1\%} \\ Harmless: \textcolor{blue}{#2\%}} \vspace{0.3em}}

\begin{table*}[!t]
    \centering
    \caption{\textbf{Most differentiating habits underlying LLMs' harmful and harmless responses when confronted with safety-related user queries from Harmbench~\citep{mazeika2024harmbench-automated-red-teaming}}.}
    \label{tab:safety-related-habits}
    \resizebox{\textwidth}{!}{
    \begin{tabular}{M{3cm}M{2cm}M{3cm}M{3cm}M{3cm}}
    \toprule
\textbf{Model} & \textbf{\% Harmful}  & \multicolumn{3}{c}{\textbf{Evidently Differentiating Habits (Top 3)}}  \\ \midrule
    \multirow{2}{*}{\vspace{-1em}\llm{DeepSeek-R1}}  & \multirow{2}{*}{\vspace{-1em}30/200}  & Listening with Understanding and Empathy   & Thinking about Thinking & Taking Responsible Risks  \\ \cline{3-5}
         &   & \hhratio{80.8}{3.3} & \hhratio{72.5}{33.3} & \hhratio{66.7}{34.1} \\ \hline
    \multirow{2}{*}{\vspace{-1em}\llm{Qwen3-32B}}     & \multirow{2}{*}{\vspace{-1em}53/200} & Taking Responsible Risks & Applying Past Knowledge to New Situations   & Creating, Imagining, and Innovating \\ \cline{3-5}
         &  & \hhratio{47.2}{19.3}   & \hhratio{40.6}{17.2} & \hhratio{46.2}{20.0} \\ \hline
    \multirow{2}{*}{\vspace{-1em}\llm{Qwen3-235B-A22B}}     & \multirow{2}{*}{\vspace{-1em}40/200} & Creating, Imagining, and Innovating  & Applying Past Knowledge to New Situations & Persisting \\ \cline{3-5}
         &  &  \hhratio{47.5}{16.7}  & \hhratio{41.2}{16.3} & \hhratio{42.5}{18.9} \\ \hline
    \multirow{2}{*}{\vspace{-1em}\llm{QwQ-32B}}     & \multirow{2}{*}{\vspace{-1em}54/200} & Listening with Understanding and Empathy  &  Applying Past Knowledge to New Situations  & Taking Responsible Risks \\ \cline{3-5}
         &    & \hhratio{63.1}{18.5} & \hhratio{60.2}{27.0} & \hhratio{52.8}{28.4} \\ \hline
    \multirow{2}{*}{\vspace{-1em}\llm{Doubao-1.5-thinking}}     & \multirow{2}{*}{\vspace{-1em}44/200} & Creating, Imagining, and Innovating   & Taking Responsible Risks & Persisting \\ \cline{3-5}
         &  &  \hhratio{46.6}{8.5}  & \hhratio{42.0}{8.5} & \hhratio{30.7}{6.5} \\ 
    \bottomrule
    \end{tabular}
    }
\end{table*}

\noindent \textbf{Experimental Setup}.  
We utilize 200 safety-related user queries from HarmBench (standard behavior subset)~\citep{mazeika2024harmbench-automated-red-teaming}. 
The experimental procedure retains the \textbf{CoT Observation} and \textbf{Habit Extraction} steps from \ourbench. 
For each task, we independently assess the presence of all 16 candidate cognitive habits within the reasoning CoTs.
To identify harmful responses, we adopt the official LLM classifier provided by~\citet{mazeika2024harmbench-automated-red-teaming}. 
Given that some models, such as \llm{o4-mini} and \llm{Claude-3.7-sonnet}, produce very few harmful responses, we exclude them from our analysis. 
We focus instead on five representative LRMs: \llm{DeepSeek-R1}, \llm{Qwen3-32B}, \llm{Qwen3-235B-A22B}, \llm{QwQ-32B}, and \llm{Doubao-1.5-thinking-pro}.
Our goal is to compare the cognitive habits underlying harmful and harmless responses. 
To ensure representativeness, we exclude any cognitive habit whose occurrence rate is below 10\% in both harmful and harmless CoTs.
This yields the final sets of evidently differentiating cognitive habits.

\noindent \textbf{Main Results}.  
Empirically, we find that specific cognitive habits are strongly associated with the generation of either harmful or harmless responses. 
As shown in~\Cref{tab:safety-related-habits}, LRMs that demonstrate strong reasoning capabilities still engage with safety-related queries. 
Notably, the most distinguishing cognitive habit is \habit{Listening with Understanding and Empathy} on \llm{DeepSeek-R1}, which appears in 80.8\% of harmful responses but only 3.3\% of harmless ones.
A broader analysis reveals that certain habits consistently correlate with harmful or harmless responses across multiple models. 
For example, the habit \habit{Taking Responsible Risks} is more frequently associated with harmful responses across all the LRMs considered. 
This pattern suggests that LRMs may be aware of the risks inherent in generating harmful responses but still choose to \habit{take responsible risks}, proceeding in fulfilling the harmful user queries.
These findings highlight the potential of utilizing cognitive habits for monitoring and mitigating the susceptibility of LLMs to external threats, thereby enhancing model safety.

\section{Related Work}
\noindent \textbf{Cognitive Behaviors of LLMs.}
As LLMs increasingly align with human capabilities~\citep{ouyang2022instructgpt,antropic-bai2022constitutional-ai}, studying their potential cognitive behaviors becomes essential. 
\citet{jones2022capturing-failures-of-llms-via-human-cognitive-biases,shaikh2024cbeval-cognitive-bias-evaluation,lin2023mind-the-biases} investigate how LLMs exhibit human-like cognitive biases, such as the framing effect.
\citet{zhang2024understanding-dark-side-of-llms-intrinsic-self-correction} demonstrates human-like cognitive traits, \eg, perfectionism in self-correction procedures.
\citet{pan2023llms-mbti-testing} show that LLMs manifest stable personality traits when evaluated using the MBTI framework.
Several studies~\citep{zeng2024johnny-persuasion-as-jailbreak,xu2024earth-flat-because} show that adversarial prompts leveraging principles from cognitive science can increase LLM compliance with harmful queries.
\citet{li2025ai-awareness} study the cognitive awareness of LLMs in a broad context.
\citet{xu2024llm-course-correction} demonstrate that LLMs can reflect their potential mistakes during response generation in a human-like manner.
Similarly, \citet{gandhi2025cognitive-behaviors-enable-rl-training-of-thinking-models} propose integrating training data associated with cognitive behaviors during the cold-start phase of reasoning reinforcement learning.
This work, unlike prior studies, identifies the reasoning CoTs in LRMs as a new opportunity and systematically investigates their cognitive habits, which are patterns that consistently emerge across diverse tasks rather than appearing only in rare cases.

\noindent \textbf{Evaluating Large Reasoning Models.}
Similar to the revealed distinction in cognitive habits by this work, recent studies have begun to distinguish LRMs from traditional non-reasoning LLMs~\citep{xu2025towards-lrm-survey,li2025system-1-to-system-2-survey-lrm}.
\citet{yue2025does-rl-incentivize-reasoning-capacity} examines the empirical upper bound of LRMs’ reasoning capabilities.
A notable phenomenon is the tendency of LRMs to overthink~\citep{chen2024not-tencent-llm-overthinking,sui2025stop-overthinking-efficient-reasoning-for-lrms-survey}, repeatedly engaging in problem-solving without regard for efficiency or cost.
This behavior can be viewed as a drawback of their capacity to arbitrarily \habit{think about thinking} and \habit{think flexibly}, underscoring the necessity of systematically investigating the cognitive tendencies of LRMs.
In addition, \citet{zhang2025should-safety-of-lrms} address the safety aspects of LRMs, identifying key components for secure deployment of LRMs.
\citet{sun2025detection-hallucination-of-lrms} further explore hallucination issues, highlighting a critical aspect of LRM reliability.
In contrast to these works, our study takes a principled behavioral perspective, systematically analyzing LRMs through the lens of human cognitive habits.

\section{Discussions}
\label{subsec:lrm-resemblance}
\noindent \textbf{Why do certain inter-family models exhibit similar cognitive habit profiles?}
We hypothesize that this resemblance can be attributed to two possible factors:
(1) \textbf{Technical Similarity in Training Methodologies}: Both \llm{Qwen-3} and \llm{DeepSeek-R1} are trained using GRPO-based reasoning RL~\citep{shao2024deepseekmath-grpo}, possibly with comparable training data distributions.
As a result, different models may independently converge to similar cognitive patterns, which likely reflect the cognitive habits studied in this work as fundamental strategies for solving complex tasks.
(2) \textbf{Indirect Data Contamination During Pre-Training}:
Operationally, since LLMs are typically pre-trained on large-scale corpora collected from the Internet, their training data may be indirectly influenced by earlier released models, even if unintentionally. 
This may explain the strikingly analogous CoTs observed in some cases, as illustrated in~\Cref{fig:qualitative-example-of-similar-cot}.
Confirming both hypotheses requires access to the models’ training details. 
We leave a deeper analysis of these similarities for future work.

\noindent \textbf{Implications of Measuring Cognitive Habits of LRMs}.
Understanding and monitoring the cognitive habits of LRMs offer several important implications.
First, it provides a window into model generalization strategies, revealing how models internalize problem-solving heuristics. 
Likewise, \citet{gandhi2025cognitive-behaviors-enable-rl-training-of-thinking-models} demonstrate that incorporating such heuristics into the cold-start stage can considerably boost reasoning performance.
Second, it enables diagnostic tools for model auditing and interoperability.
CoTs can help developers and users trace the underlying reasoning behind model responses, thereby supporting transparency and accountability~\citep{openai2025misbehavior-cot-monitoring}.
Lastly, these insights can guide the design of training objectives and data curation practices to encourage diversity and reduce unintended behavioral convergence.
This is particularly relevant given that similar cognitive habits across models may reflect shared inductive biases or potential training data leakage.

\vspace{1em}
\section{Conclusion and Future Work}

We investigate whether Large Reasoning Models exhibit cognitive habits.
To advance this, we adapt the \textit{Habits of Mind} framework to develop \ourbench, a benchmark tailored for cognitive habit evaluation.
\ourbench is designed to satisfy key principles: habit specificity, spontaneity, real-world utility, comprehensiveness, and scalability.
Based on the characteristics of each habit, we adopt a hybrid task construction approach:
For reasoning-related habits, we incorporate math problems from existing academic datasets; for others, we define task-generation principles and leverage advanced LLMs to generate tasks at scale.
We then employ an evidence-first extraction method that identifies the presence of target cognitive habits in LRMs' CoTs.
Using \ourbench, we conduct a comprehensive evaluation of 16 widely recognized LLMs.
Our results reveal diverse patterns in cognitive habit profiles, with LRMs showing clear advantages over conventional LLMs in exhibiting such habits.
The findings confirm that LRMs do display human-like cognitive habits and uncover intriguing relationships across model families, including intra-family, post-distillation, and inter-family similarities.
Our extension to safety-related instructions demonstrates the potential of monitoring CoT behavioral patterns to detect LRM misbehavior.
We expect this work to inspire future efforts in interpreting and auditing LLM behavior across diverse applications.

In this study, the identification of cognitive habits is based on the detection of meta-thinking statements within the CoT. 
While this method offers interpretability and precision, it may overlook implicit cognitive habits that influence reasoning but are not directly articulated. 
Future work could focus on identifying such hidden cognitive habits through more advanced techniques.
Additionally, some habits—such as \habit{Gathering Data Through All Senses}—are inherently linked to more open or perceptual settings, such as multimodal reasoning~\citep{liu2024llava,bai2025qwen2-5-vl-technical-report}. 
As most current LRMs primarily support text-only inputs and outputs, we plan to extend our benchmark to evaluate future LRMs with multimodal capabilities.

\clearpage

\bibliography{custom}

\clearpage
\appendix

\end{document}